\documentclass[letterpaper, 10 pt, journal, twoside]{IEEEtran}

\usepackage{times}
\usepackage{epsfig}
\usepackage{graphicx}
\usepackage{amsmath}
\usepackage{amssymb}

\usepackage{color}
\usepackage[nolist, nohyperlinks]{acronym}
\usepackage{siunitx}
\usepackage{pifont}% http://ctan.org/pkg/pifont
\usepackage{flushend}

\newcommand{\br}{\mathbf{r}}

\newcommand{\figref}[1]{Fig.~\ref{#1}}
\newcommand{\secref}[1]{Section~\ref{#1}}

\newcommand{\eqnref}[1]{Eq.~\eqref{#1}}
\newcommand{\tabref}[1]{Table~\ref{#1}}

\def\mb{\mathbf}

\newcommand{\tf}[2]{{}^{#1}{\mathbf{T}_{#2}}}
\newcommand{\tfl}[2]{{}^{#1}{\mathbf{K}_{#2}}}

\makeatletter
\DeclareRobustCommand\onedot{\futurelet\@let@token\@onedot}
\def\@onedot{\ifx\@let@token.\else.\null\fi\xspace}

\makeatother

\newcommand{\boldparagraph}[1]{\vspace{0.2cm}\noindent{\bf #1:} }

\definecolor{darkgreen}{rgb}{0,0.7,0}

\newcommand{\methodName}{\textsc{Star-Track}}
\newcommand{\cmark}{\ding{51}}%
\newcommand{\xmark}{\ding{55}}%

\definecolor{ellisred}{rgb}{0.87,0.44,0.38} % 222, 112, 97
\definecolor{ellisgreen}{rgb}{0.69,0.90,0.52} % 176, 230, 133
\definecolor{elliscyan}{rgb}{0.29,0.77,0.74} % 74, 196, 189
\definecolor{ellisorange}{rgb}{0.89,0.55,0.28} % 227, 140, 71
\definecolor{ellisblue}{rgb}{0.41,0.61,0.86} % 105, 156, 219

\begin{acronym}
    \acro{mot}[MOT]{\textit{multi-object tracking}}
    \acro{lmm}[\textsc{lmm}]{\textit{latent motion model}}
    \acro{amota}[AMOTA]{\textit{average multi object tracking accuracy}}
    \acro{amotp}[AMOTP]{\textit{average multi object tracking precision}}
    \acro{ids}[IDS]{\textit{number of identity switches}}
    \acro{frag}[FRAG]{\textit{number of track fragmentations}}
    \acro{mt}[MT]{\textit{number of mostly tracked trajectories}}
    \acro{ml}[ML]{\textit{number of mostly lost trajectories}}
    \acro{sota}[SOTA]{\textit{state-of-the-art}}
    \acro{nms}[NMS]{\textit{non-maximum suppression}}
    \acro{tfnet}[TfNet]{\textit{transformation hyper-network}}
    \acro{fpn}[FPN]{\textit{feature pyramid network}}
    \acro{ffn}[FFN]{\textit{feed forward network}}
    \acro{3d}[3D]{3D}
    \acro{mln}[MLN]{motion-aware layer normalization}
\end{acronym}

\begin{document}

\title{S.T.A.R.-Track: Latent Motion Models for End-to-End 3D Object Tracking with Adaptive Spatio-Temporal Appearance Representations}
\author{Simon Doll$^{1,2}$, Niklas Hanselmann$^{1,2}$, Lukas Schneider$^{1}$, Richard Schulz$^{1}$, \\ Markus Enzweiler$^{3}$, Hendrik P.A. Lensch$^{2}$% 

\thanks{Manuscript received: 7, 26, 2023; Revised 10, 9, 2023; Accepted 12, 5, 2023.}%Use only for final RAL version
\thanks{This paper was recommended for publication by Editor Ashis Banerjee upon evaluation of the Associate Editor and Reviewers' comments.
This work was supported by the German Federal Ministry for Economic Affairs and Climate Action (KI Delta Learning: Förderkennzeichen 19A19013A). } %Use only for final RAL version
\thanks{$^{1}$Mercedes-Benz AG {\tt \footnotesize simon.doll@mercedes-benz.com}}%
\thanks{$^{2}$University of Tübingen}%
\thanks{$^{3}$Institute for Intelligent Systems, Esslingen University of Applied Sciences}        
\thanks{Digital Object Identifier (DOI): 10.1109/LRA.2023.3342552}
}

\markboth{IEEE Robotics and Automation Letters. Preprint Version. Accepted 12, 2023}
{Doll \MakeLowercase{\textit{et al.}}: S.T.A.R.-Track}

\maketitle

\begin{abstract}
    Following the tracking-by-attention paradigm, this paper introduces an object-centric, transformer-based framework for tracking in 3D. 
    Traditional model-based tracking approaches incorporate the geometric effect of object- and ego motion between frames with a geometric motion model. Inspired by this, we propose \methodName\, which uses a novel \ac{lmm} to additionally adjust object queries  to account for changes in viewing direction and lighting conditions directly in the latent space, while still modeling the geometric motion explicitly. Combined with a novel learnable track embedding that aids in modeling the existence probability of tracks, this results in a generic tracking framework that can be integrated with any query-based detector.
    Extensive experiments on the nuScenes benchmark demonstrate the benefits of our approach, showing \ac{sota} performance for DETR3D-based trackers while drastically reducing the number of identity switches of tracks at the same time.
\end{abstract}
\begin{IEEEkeywords}
  % IROS
  % Autonomous Vehicle Navigation, AI-Based Methods, Deep Learning for Visual Perception
  Visual Tracking, Deep Learning for Visual Perception, Autonomous Vehicle Navigation
  %RAL Deep Learning Methods, Deep Learning for Visual Perception, Visual Tracking.
\end{IEEEkeywords}

\IEEEpeerreviewmaketitle

\section{Introduction}
\IEEEPARstart{R}{obust} perception and tracking of movable objects in the environment form the basis for safe decision-making in autonomous agents such as self-driving cars.
Classical \ac{mot} pipelines follow a \emph{tracking-by-detection} paradigm, using object detectors coupled with greedy matching~\cite{li2022unifying} and state estimators~\cite{badue2021self, ess2010object} to track objects.
Building on recent advances in object detection from multi-view camera images, transformer-based architectures~\cite{wang2022detr3d, doll2022spatialdetr, wang2022focal} can yield strong tracking performance~\cite{li2022unifying, nuscenes_tracking_benchmark} using relatively low-cost sensors. 
However, decoupling the detection and tracking tasks comes with two main drawbacks: (1) the object detection model is optimized towards a detection metric, rather than directly optimizing for the downstream tracking performance, which is prone to compounding errors~\cite{karkus2022diffstack,gu2023vip3d} and (2) it makes it non-trivial to incorporate appearance information, which poses a challenge to consistent association. 
This in particular can lead to difficulties in handling confusion among object identities in crowded scenarios with many partial object-to-object occlusions~\cite{meinhardt2022trackformer}.
\begin{figure}[t]
\begin{center}
   \includegraphics[width=1\linewidth]{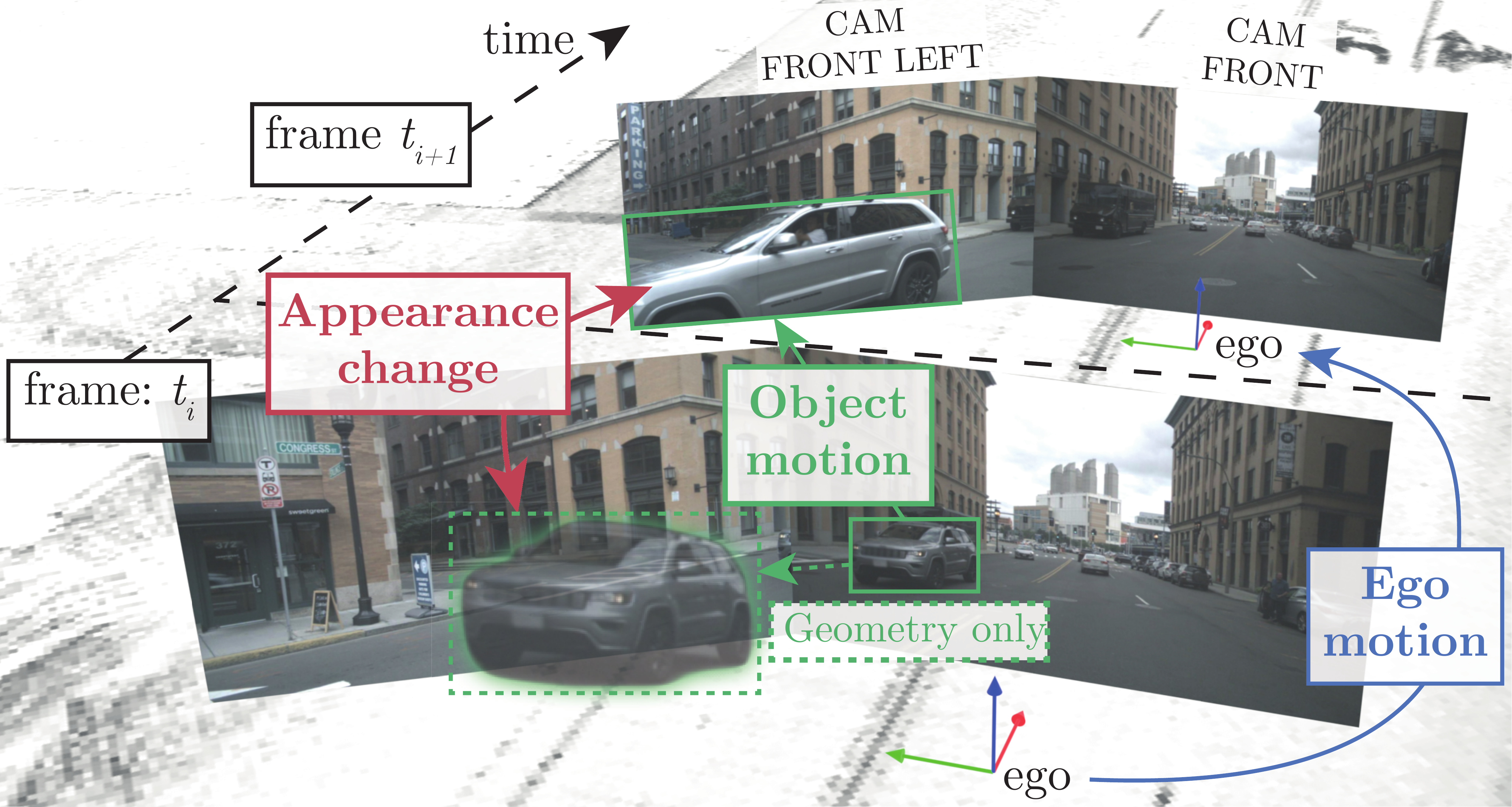}
\end{center}
   \caption{Visualization of a tracked object for two consecutive frames. Due to ego and object motion the 3D pose and the appearance of the object in the camera images change in scale, viewing angle and lighting condition. We utilize an explicit geometric and a novel latent motion model to compensate for these effects during the prediction step of the tracking pipeline.}
\label{fig:teaser}
\end{figure}

Recent works~\cite{zeng2022motr, zhang2022mutr3d} propose an alternative \emph{tracking-by-attention} paradigm that unifies perception and tracking into a single module. 
Under this paradigm, the geometric and semantic information contained in the latent object queries of query-based detectors can be leveraged for the association of object instances across time via attention~\cite{vaswani2017attention}. 
Additionally, tracking-by-attention allows to use these queries as detection priors in the following frames. 
This requires adjusting them to the expected future object state, analogous to the model-based prediction step in classical state estimator-based trackers~\cite{pang2023simpletrack}. 

For geometric features, this can be done by applying the transformation corresponding to both ego and estimated object motion. 
However, this is not possible for latent object queries, as they also encode semantics and appearance in addition to geometric information. 
MUTR3D~\cite{zhang2022mutr3d} sidesteps this issue by anchoring object queries to geometric reference points which can be analytically updated. 
While this enables some adjustment, only the object translation rather than the full pose is considered and the change in appearance resulting from changes in the relative pose is not modeled. 
A tracking method that corrects both geometric and appearance information directly in latent space is proposed in~\cite{ruppel2022transformers}. However, this approach forfeits the ability to analytically update geometric information and does not model object motion.

In this paper, we propose to compensate for appearance changes resulting from both ego and object motion via a novel \ac{lmm} which updates queries in latent space as a function of the geometric motion.
Paired with analytical updates on geometric reference points for each query, we obtain a transformable \textbf{S}patio-\textbf{T}emporal geometry and \textbf{A}ppearance \textbf{R}epresentation for each object that enhances consistency with future observations.
Furthermore, we propose track embeddings that encode information on the lifetime of a tracked object to distinguish track queries from new detections. Our approach termed \methodName, exhibits improved tracking performance. Specifically, we observe that accounting for appearance changes between frames as well as the improved existence probability modeling eases association, leading to a drastically reduced number of switches in object instance identities.

In summary, we make the following main contributions:
\begin{itemize}
    \item We are the first to compensate for the appearance change induced by both, ego- and object motion in a tracking-by-attention paradigm leveraging a \acf{lmm} that extends query-based object detectors.
    % which can easily be applied to various kinds of query-based object detectors.
    \item We introduce novel track embeddings allowing to implicitly model the life cycle of a tracked object.
    \item We outperform current state of the art DETR3D-based tracking approaches on nuScenes~\cite{caesar2020nuscenes} where the \ac{lmm} and track embeddings in particular reduce fragmentations and identity switches by a large margin.
\end{itemize}

\section{Related Work}
\boldparagraph{Query-based Detection}
\ac{mot} approaches that follow the tracking-by-detection~\cite{li2022unifying, badue2021self, ess2010object} paradigm require a detector to detect a set of objects in each frame.
The pioneering work DETR~\cite{carion2020end} proposed a way to leverage the transformer architecture for object detection. In contrast to previous approaches, this set-based architecture comes with various desirable properties such as a sparse prediction scheme, a dynamic amount of object hypotheses, and no need for hand-crafted components such as \ac{nms}.
Additionally, the concept was generalized to the 3D case as well as to different sensor modalities including LiDAR~\cite{bai2022transfusion, erabati2023li3detr}, multi-view camera~\cite{wang2022detr3d, doll2022spatialdetr} and multi-modal detection methods~\cite{li2022unifying, bai2022transfusion}. It is noteworthy that such query-based detectors became the de-facto standard in object detection and reach \ac{sota} performance on various benchmarks such as COCO~\cite{lin2014microsoft} or nuScenes~\cite{caesar2020nuscenes}.

\boldparagraph{Tracking-by-Detection}
Methods that rely on the well-established tracking-by-detection paradigm have the benefit of being compatible with any detection framework since the detection per frame and the tracking/association part are not directly linked. 
A simple greedy association~\cite{yin2021center} is still widely adopted in \ac{sota} methods on the nuScenes tracking benchmark~\cite{li2022unifying, nuscenes_tracking_benchmark, bai2022transfusion}.
In this generic approach, the detector can not make use of previous tracks and the association often relies on geometric cues only. This causes track identity switches in which a track is reinitialized with a detection instead of the detection being associated with the previous track.
Various extensions such as re-ID features~\cite{pang2021quasi},~\cite{ristani2018features} and motion models~\cite{ess2010object, schubert2008comparison} have been proposed to mitigate this effect.
Motion models integrate prior knowledge about the physical properties and trajectory of the tracked object while re-ID features allow an association that is not solely based on bounding box geometry but also influenced by other features such as motion cues or objects appearance.
\begin{figure*}[t!]%TODO double check placement
  \includegraphics[width=\textwidth]{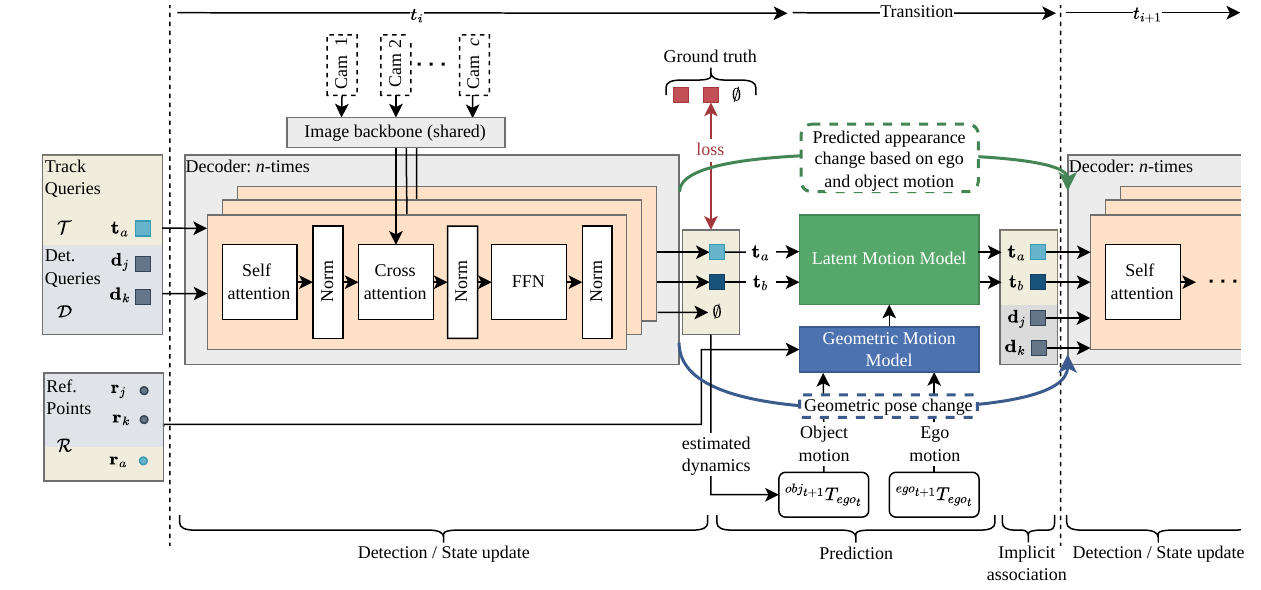}
  \caption{\methodName\ architecture. A joint set of time-independent object queries and track queries of the previous frames is used in a stack of decoder layers that utilize self- and cross-attention blocks to detect and re-identify objects in consecutive time steps. This requires predicting the state of each object in the following frame. Combined with any geometric motion model (blue) the newly proposed latent motion model (green) solves this issue by modeling the spatio-temporal change of a track query in the latent and the 3D geometric space jointly, based on the estimated dynamics.
  }
  \label{fig:architecture}
\end{figure*}

\boldparagraph{Tracking-by-Attention}
To overcome the independent nature of the detection and tracking modules in a fully differentiable fashion and to implicitly solve the association between frames, the \textit{tracking-by-attention} paradigm can be used~\cite{meinhardt2022trackformer, sun2020transtrack}.
Leveraging the potential of attention, tracking and detection are performed jointly by auto-regressive query-based tracking since each detection of the last frame is used as a prior (track-query) for the next frame.
MUTR3D\cite{zhang2022mutr3d} extends the object detection method DETR3D~\cite{wang2022detr3d} for tracking by adding a geometric compensation of object and ego motion.
This is done by utilizing a 3D reference point per object that is transformed between consecutive frames while the latent query features remain unchanged. 
A possibility to account for an appearance change caused by the ego-motion is presented in~\cite{ruppel2022transformers}.
The proposed ego-motion-compensation module models the effect directly in the latent space as a linear function that depends on the estimated transformation between the two time steps.
Similar to the 3D case in which the transformation can be represented as a homogeneous matrix this transformation in latent space is modeled as a full-rank matrix which is learned from the given ego motion via a hyper-network~\cite{ha2017hypernetworks}.

Inspired by the aforementioned previous works, we propose a latent motion model to account for the effects of \textit{ego and object motion} on the latent appearance representation jointly. 
This allows for keeping the explicit geometric update proposed in~\cite{zhang2022mutr3d} while altering the learned appearance of an object as a function of the geometric transformation to simplify its detection and re-identification in the next frame.

\section{Method}
Our proposed approach tackles \acl{mot} from multi-view camera images. Given a set of $c$ mono camera images $\mathbf{I}_c \in \mathcal{I}_t$ with shape $H \times W \times 3$ for each timestamp $t$ the tracking objective is to estimate a set of bounding boxes $\mathbf{b}_t^{id} \in \mathcal{B}_t$ with $\mathbf{b}_t^{id} = \begin{bmatrix} x, y, z, w, l, h, \theta, v_x, v_y \end{bmatrix}$ describing each object as defined in nuScenes~\cite{nuscenes_tracking_benchmark}. Besides center, shape, heading angle and velocity of each object, each bounding box has a corresponding $id$ that is consistent over time.
\subsection{Overall Architecture}
Under the tracking-by-attention paradigm, an object is tracked by updating its unique query feature to be consistent with new observations and other track hypotheses at each point in time via the attention mechanism~\cite{meinhardt2022trackformer, zhang2022mutr3d}.
Since attention reasons about the affinity between new observations and existing tracks via feature similarity, queries of tracked objects need to be adjusted to account for the changes in relative pose and appearance resulting from both ego- and object motion between frames.
To this end, we propose an \ac{lmm}, an extension to commonly used purely geometric motion models.
The \ac{lmm} adjusts the latent feature of each object to be consistent with the expected state in the next frame, increasing similarity to new observations of the same object and simplifying the association task.
The \ac{lmm} implements a generic query prediction strategy that can be readily coupled and jointly trained with any query-based detector.

An overview of the proposed architecture is presented in Fig.~\ref{fig:architecture}.
We utilize a decoder-only transformer architecture as in DETR3D~\cite{wang2022detr3d}, where a set of learnable detection queries $\mathcal{D}=\{\mathbf{d}_1,\dots \mathbf{d}_n\}$ is used to represent hypotheses for newly detected objects in the scene.
Following the MUTR3D~\cite{zhang2022mutr3d}, the time independent detection queries are concatenated with a set of track queries $\mathcal{T} = \{\mathbf{t}_1,\dots \mathbf{t}_m\}$ that correspond to hypotheses from the previous frames. 
Then, the decoder refines both the track hypotheses and new detections jointly by applying self- and cross-attention into features extracted from multi-view camera images $\mathcal{I}_t$ by a shared image backbone in an alternating fashion.
We kindly refer the reader to~\cite{zeng2022motr, zhang2022mutr3d, carion2020end} for further details on the general \ac{mot} architecture. 
Lastly, the bounding boxes $\mathcal{B}_t$ are obtained with a \ac{ffn} while we carry the objects over to the next frame by applying both the analytical geometric motion transformation as well as the \ac{lmm}.

\subsection{Revisiting Multi-Object Tracking}\label{sec:tracking_structure}
Model-based tracking systems~\cite{badue2021self, kalman1960new} typically rely on sequential steps that allow to incorporate inductive biases into the different parts of the tracking framework while also maintaining a high level of interpretability.

\boldparagraph{Detection / State update}    
In each frame a set $\mathcal{D}$ of new detections is used to update the current belief state of tracked objects $\mathcal{T}$ in the scene.
This enables rejecting implausible sensor measurements, updating the estimated bounding box and existence probability of each track, and spawning new tracks for newly appeared objects. The transformer-based tracking-by-attention mechanism mirrors this behavior by performing two attention operations
utilizing \textit{scaled dot product attention}~\cite{vaswani2017attention}. \textit{Self-attention} within the joint set of track queries and newly spawned detection queries models object interactions, integrating new objects and rejecting duplicate proposals.
Subsequently, \textit{cross-attention} between all object queries and the camera features is used to refine each object proposal by incorporating sensor measurements. 
The tracking-by-attention framework utilizes track queries of previous time steps as priors for the detection in the next frame which potentially simplifies the detection of objects that are far away, partially occluded, or hardly visible. 

\boldparagraph{Prediction}
Given the current ego motion transformation ${}^{ego_{t+1}}\mathbf{T}_{ego_{t}}$~\eqnref{eq:tf:ego} and estimated object dynamics, for instance the velocity of each tracked object, a traditional geometric tracking framework predicts the object pose in the next frame.
This is typically achieved utilizing a motion model which is a function of object state and dynamics.

For a latent object representation the geometric update in terms of the object pose should be handled similarly to the explicit bounding box representation since the geometric transformation can be applied analytically. 
However, the high-dimensional appearance representation of the object query also needs to be taken into consideration since the ego and object motion might heavily affect the appearance of the object and thus its query feature in the next frame, see~\figref{fig:teaser}. 
This is crucial since the transformer attention relies on a query-key similarity~\cite{vaswani2017attention}.
Without a latent appearance update the re-identification of a tracked object in the next frame might be impaired.
Firstly, track identity switches or track losses can occur if a track query cannot be associated to the sensor data of the next frame in the cross-attention blocks. 
Secondly, without proper appearance updates, duplicates might spawn, since existing tracks fail to suppress their newly detected counterparts in the self-attention blocks.

\boldparagraph{Association}
To associate detections in the next frame with existing tracks, any similarity metric between object hypotheses can be used.
Traditional methods rely on geometry-based metrics~\cite{yin2021center,luo2021multiple} or additional re-ID features~\cite{pang2021quasi, ciaparrone2020deep} to form an affinity matrix between tracks and new detections which can be used together with the Hungarian algorithm to find an optimal matching.
Auto-regressive query-based tracking methods~\cite{meinhardt2022trackformer, zeng2022motr, zhang2022mutr3d, sun2020transtrack} solve this problem differently since a track query always represents the same object in the scene resulting in an implicit association.
During training, this is enforced by matching each track query to its corresponding object in the scene to which it was assigned at first appearance. 
If two hypotheses describe the same object, the model needs to distinguish between newly spawned and already tracked objects and favor the latter.
This is crucial since confusions between tracks and newborn detections might result in track losses or identity switches between tracks and new detections at inference time.

As a result of the considerations above, two key challenges arise for auto-regressive query-based tracking: (1) The prediction step needs to model the influence of the geometric transformation on the pose of the object as well as its latent appearance and semantic features.
(2) Due to the implicit association mechanism each track query needs a latent existence probability to efficiently suppress newborn duplicate queries that also belong to the tracked object.

\subsection{Latent Motion Models}\label{sec:lmm}
The prediction step in the tracking pipeline aims to estimate the state of an object in the next timestep.
In our model, the set of tracked objects and newly spawned detections is defined as a set of latent vectors $\mathbf{q} \in \mathcal{Q}$.
Additionally, the position of each object query is defined with respect to a 3D reference point $\mathbf{r} \in \mathcal{R}$ as proposed in~\cite{wang2022detr3d}.
Consequentially, the geometric effect of the ego motion for a time delta $\delta_{t}$ between two frames can be described with a homogeneous matrix that combines rotation $\mathbf{R}$ and translation $\mathbf{t}$
\begin{equation}
{}^{{ego}_{t+1}}\mathbf{T}_{ego_t} 
    = \begin{bmatrix}\mathbf{R} & \mathbf{t} \\
		0          & 1
	\end{bmatrix}.
 \label{eq:tf:ego}
\end{equation}

Furthermore, the regression branches of the transformer decoder predict an estimate of the dynamics for each object. These include the estimated velocity $\mathbf{v} = \begin{pmatrix} v_x & v_y \end{pmatrix}$, that is supervised by ground truth data during training, and an optional turn-rate $\delta_\theta$ for the heading angle $\theta$ resulting in
\begin{equation}
 \tf{e'_t}{e_t}
 % &=\tf{{o}_{t+1}}{e_t} \\
 = \begin{bmatrix}
	   \cos(\delta_\theta) & -\sin(\delta_\theta) & 0 & v_x \cdot \delta_{t}\\
          \sin(\delta_\theta) & \cos(\delta_\theta) & 0  & v_y \cdot \delta_{t}\\
          0                   & 0                   & 1  & 0  \\
          0                   & 0                   & 0  & 1
	\end{bmatrix}.
\end{equation}
For consistent notation, we propose an auxiliary frame $e'_t$ that describes the state of the world after object motion compensation relative to the ego frame at time $t$. 
We note that due to the explicit modeling of this transformation, any motion model~\cite{luo2021multiple} can be used to constrain the estimated transformations by model-based assumptions.

\boldparagraph{Hyper-Networks}
Besides the explicit geometric update on the reference point $\br$ of an object as defined in \eqnref{eq:geometric_motion_compensation}, an additional update to the latent features $\mathbf{q}$ is required to propagate those to the next frame. As argued in~\cite{ruppel2022transformers}, the effect of the geometric transformation in latent space can by modeled as a linear operator that performs an input-dependent multiplication on the object query in the form of a latent transformation matrix $\tfl{b}{a}$.
This matrix is a function of its geometric counterpart $\tf{b}{a}$ and represents an arbitrary transformation from frame $a$ to frame $b$. Geometric and latent information is jointly updated: 
\begin{eqnarray}
    \mb{r}_{e_{t+1}} &= \tf{e_{t+1}}{e'_t} \cdot \tf{e'_t}{e_t} \cdot \mb{r}_{e_t} & \text{Geometric Update} \label{eq:geometric_motion_compensation} \\
    \mb{q}_{e_{t+1}} &= \tfl{e_{t+1}}{e'_t} \cdot \tfl{e'_t}{e_t} \cdot \mb{q}_{e_t} & \text{Latent Update} \label{eq:latent_motion_compensation}
\end{eqnarray}

\begin{figure}[t]
  \includegraphics[width=\linewidth]{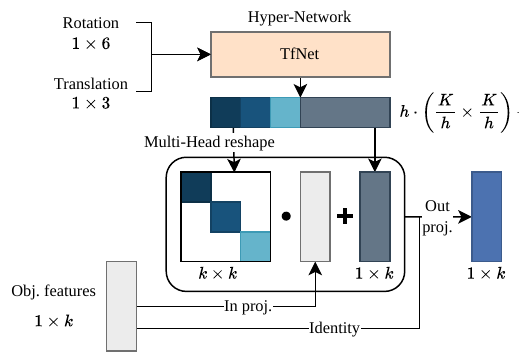}
  \caption{Latent motion model architecture. A geometric transformation consisting of a translation and rotation is applied to the high-dimensional object query by using a sparse latent transformation matrix $K$. We estimate the elements of $K$ with a hyper-network (TfNet) and apply the transformation as an input dependent multiplication, mimicking the behavior of a homogeneous matrix in 3D. Note the sparse block-diagonal shape of the generated matrix.}
  \label{fig:lmm_architecture}
\end{figure}

We propose a \ac{tfnet} to estimate the parameters of the latent transformation matrix $\tfl{b}{a}$.
This matrix is applied as an input-dependent multiplication with the latent object query $\mathbf{q}$.
A latent translational offset is incorporated as an element-wise addition. An overview of the proposed \ac{lmm} architecture is given in \figref{fig:lmm_architecture}. 

\boldparagraph{Input Representation}
The input to the \ac{tfnet} consists of a rotational and translational part:
\begin{equation}
    \tfl{b}{a} = \text{TfNet}({}^b\mathbf{R}_a, {}^b\mathbf{t}_a),
\end{equation}
whereas ${}^a\mathbf{R}_b$ describes the rotational component and ${}^a\mathbf{t}_b$ the translation of ${}^{{a}}\mathbf{T}_{b}$. While the translation is represented as a 3D vector, we utilize the 6D rotation representation proposed in~\cite{zhou2019continuity} to increase the numeric stability.

\boldparagraph{Sparse Latent Transforms}Since the latent features are typically high-dimensional~\cite{wang2022detr3d, zhang2022mutr3d}, a hyper-network that predicts $\tfl{b}{a}$ as a full-rank matrix might be over-parameterized or even intractable to train.
This is due to the  large number of parameters in the output weight matrix $\tfl{b}{a}$ that need to be computed per object in each frame.
We mitigate this potential issue by adopting the concept of multi-head attention from~\cite{wang2022detr3d, vaswani2017attention,  carion2020end} and propose a sparse multi-head \ac{lmm}.
Here, attention is computed as a combination of $h$ different low-dimensional attention heads that operate on $h$ splits of the feature vector with a dimensionality of $h_{dim} = k / h$ each.
Instead of predicting $k^2 = h^2 \cdot h_{dim}^2$ weights for a full-rank description of $K$, we propose to only predict $h \cdot h_{dim}^2$ weights for a sparse approximation that drastically reduces the parameter count of the latent transformation matrix. Analogously to the attention computation, these are then used as heads along the diagonal of $\tfl{b}{a}$ that operate on parts of the $k$-dimensional latent vector $\mb{q}$, see~\figref{fig:lmm_architecture}. Since only neighboring dimensions of the feature vector that lie within the same head can influence the latent transform, we follow~\cite{vaswani2017attention} and incorporate an input and output projection to mitigate this effect.

As a result, with the multi-head \ac{lmm} the latent transformation can be directly applied to the full latent vector in a sparse and numerically more stable fashion, while also streamlining the architecture to follow the layout of the attention blocks that are used in all other parts of the model.

\subsection{Track Embeddings for implicit Existence Probability} % Modeling}
As discussed in \secref{sec:tracking_structure}, the self-attention blocks serve the purpose of allowing for object interactions as well as suppressing newborn detections that belong to an already tracked object. Although it might be sufficient to distinguish between tracks and new detections in this case, the track queries in general require a consistent integration of the track history to account for short-term occlusions and deliver robust existence probability estimates.

Since learned embeddings have been used successfully to incorporate inductive biases in attention-based detectors~\cite{bai2022transfusion,doll2022spatialdetr}, we propose to use a learned latent \textit{track embedding} to address the aforementioned issues. Using a single shared track embedding $\mathbf{e}$ and a \ac{ffn} we update all active tracks $\mathcal{T}$ of the current time step using
\begin{equation}
    \mathbf{t_i}' =  \mathbf{t_i} + \text{FFN}([\mathbf{t_i}, \mathbf{e}]) \hspace{12pt} \forall \mathbf{t_i} \in \mathcal{T}.
\end{equation}
This way, the model is flexible to integrate the track embedding to the current latent state of an object and to model the desired distinction between tracks and new detections. As a result, we obtain more consistent existence probabilities and improved track losses, track fragmentations and identity switches, as our experiments in \secref{subsec:experiments_main_comparison} show. 
\section{Experiments}
We evaluate the performance of \methodName\ on the tracking task~\cite{nuscenes_tracking_benchmark} of the nuScenes dataset~\cite{caesar2020nuscenes}.
Additionally, we provide extensive ablation studies to evaluate the effects of different \ac{lmm} configurations, latent track embeddings and transform representations, as well as qualitative results.

\subsection{Experimental Setup}\label{sec:exp_setup}

\boldparagraph{Dataset}
All experiments are performed on the large-scale nuScenes dataset~\cite{caesar2020nuscenes} that consists of 1000 scenes with a length of \SI{20}{\second} with a frequency of \SI{2}{\hertz}.
We use the official training, validation and test set split and the seven object classes required for the tracking benchmark~\cite{nuscenes_tracking_benchmark}.

\boldparagraph{Metrics}
We report performance using the metrics as defined in the {nuScenes} benchmark~\cite{nuscenes_tracking_benchmark}: These include the \ac{amota} as well as the \ac{amotp}. Additionally, we report the \ac{ids}, \ac{frag} and \ac{mt} as secondary metrics. For the full metric definitions and further details, we refer to~\cite{nuscenes_tracking_benchmark, caesar2020nuscenes}.

\boldparagraph{Training Configuration}
To increase comparability and reproducibility, we closely follow the settings proposed in MUTR3D~\cite{zhang2022mutr3d}. Each training sample consists of three consecutive frames. The geometric and latent motion models assume a constant velocity and no turn-rate transformation for each object, as used in~\cite{zhang2022mutr3d}. We leave the integration of more complex dynamics models to future work. As in previous works~\cite{wang2022detr3d, zhang2022mutr3d}, bi-partite matching and the Hungarian algorithm are used to match tracked objects of the current frame with the ground truth. We use Focal-Loss~\cite{lin2017focal} as classification loss and L1-Loss for bounding box regression. In the training phase, previously matched track queries are always matched to their corresponding ground truth objects. As in~\cite{zeng2022motr,zhang2022mutr3d}, we drop tracked queries with a probability {$p_{drop}=0.1$} and spawn false positive tracks with a probability of \mbox{$p_{fp} = 0.3$}. During inference, non-confirmed tracks are kept as inactive for five frames to handle full occlusions over multiple time steps.

We train all models for 24 epochs with the same random seed on four NVIDIA-V100 GPUs using a batch size of four and a ResNet-101 backbone~\cite{he2016deep}. As proposed in~\cite{zhang2022mutr3d}, the transformer utilizes $l=6$ decoder layers, $q=300$ detection queries for each frame and a latent dimension of $d_l = 256$ spread over $h=8$ heads of dimension $d_h = d_l / h = 32$. This is also used as configuration of the proposed \ac{lmm}. All experiments use the training schedule proposed in DETR3D~\cite{wang2022detr3d} that utilizes a learning rate of $2e^{-4}$, a cosine annealing learning rate schedule and AdamW~\cite{loshchilov2019decoupled}.

\begin{table*}[t]
  \begin{center}
    \caption{Comparison of state-of-the-art methods on the nuScenes benchmark.
    For a fair comparison all methods on the validation set utilize DETR3D~\cite{wang2022detr3d} as detector.
    DETR3D$\dagger$ utilizes the greedy tracking approach proposed in~\cite{yin2021center}, DETR3D$\ddagger$ the more elaborate tracking approach introduced in~\cite{pang2022simpletrack}. Due to a potential evaluation error in MUTR3D~\cite{zhang2022mutr3d, mutr3d_eval_bug} we add a customized MUTR3D$^+$ baseline. The version of our model that only uses the \ac{lmm} and no learned track embedding is denoted by $^*$.} 
    % TODO add asterix for concurrent
    \label{table:nusc_results}
    \begin{tabular}{lllcccccccccc}
      \hline\noalign{\smallskip}
      Name & Backbone & $\#$Params & AMOTA$\uparrow$ & AMOTP$\downarrow$ & RECALL$\uparrow$ & MOTA$\uparrow$  & MT$\uparrow$ & FRAG$\downarrow$ & IDS$\downarrow$ & FPS$\uparrow$\\
      \noalign{\smallskip}
      \hline
      Validation-Split\\
      \hline
      DETR3D~\cite{wang2022detr3d}$\dagger$ & ResNet101 & - & 0.327 & 1.372 & 0.463 & 0.291 & 2039 & 2372 & 2712 & -\\
      DETR3D~\cite{wang2022detr3d}$\ddagger$ & ResNet101 & - & 0.353 & 1.382 & 0.469 & 0.315 & 2065 & 2309 & 1807 & -\\
      MUTR3D~\cite{zhang2022mutr3d} & ResNet101 & 59M & 0.294 & 1.498 & 0.427 & 0.267 & - & - & 3822 & 6.98\\
      MUTR3D~\cite{zhang2022mutr3d}$^+$ & ResNet101 & 59M & 0.360 & 1.411 & 0.487 & 0.341 & 2368 & 1232 & 522 & \textbf{6.98}\\
      CC-3DT~\cite{fischer2022cc} & ResNet101  & - & 0.359 & 1.361 & 0.498 & 0.326 & - & - & 2152 & 2.06\\
      PF-Track~\cite{pang2023standing} & VovNet-V2-99 & - & 0.362 & 1.363 & - & - & - & - & \textbf{300} & -\\
      \textbf{\methodName$^*$} & ResNet101 & 62M & 0.378 & 1.365 & 0.497 & 0.354 & 2467 & 1241 & 439 & 6.86\\
      \textbf{\methodName} & ResNet101 & 62M & \textbf{0.379} & \textbf{1.358} & \textbf{0.501} & \textbf{0.360} & \textbf{2468} & \textbf{1109} & 372 & 6.76\\
      \hline
      Test-Split\\
      \hline
      MUTR3D~\cite{zhang2022mutr3d} & ResNet101 & 59M & 0.270 & 1.494 & 0.411 & 0.245 & 2221 & 2749 & 6018 & -\\
      \textbf{\methodName} & VovNet-V2-99 & 83M & \textbf{0.439} & 1.256 & \textbf{0.562} & \textbf{0.406} & \textbf{3726} & 1250 & 607 & 6.68\\
      \hline
    \end{tabular}
  \end{center}
\end{table*}

We initialize the model with an already trained MUTR3D checkpoint to avoid retraining and keep the image backbone and \ac{fpn} fixed. To initialize the newly introduced \ac{lmm}, we propose a simple yet effective pretraining scheme: For each sample in the dataset we store the tracking results, consisting of latent queries as well as decoded object proposals from MUTR3D~\cite{zhang2022mutr3d} and train the \ac{lmm} to predict the state of the latent object query vectors of the next frame.

\subsection{Comparison to Existing Works}
\label{subsec:experiments_main_comparison}
We compare \methodName\ to state-of-the-art methods for 3D \ac{mot} on multi-view camera images. To control for the effects of different detection algorithms on the overall tracking performance, we present our main comparison in terms of DETR3D-based frameworks, which are well-established and widely used~\cite{doll2022spatialdetr,liu2022petr,wang2022focal}. This allows for a fair assessment of our contributions.

As shown in~\tabref{table:nusc_results}, our tracking framework \methodName\ that utilizes the novel \ac{lmm} and track embedding achieves the best performance in all key metrics on the nuScenes benchmark~\cite{nuscenes_tracking_benchmark} for DETR3D-based~\cite{wang2022detr3d} tracking algorithms without reducing the inference speed.

In comparison to the greedy tracking DETR3D baseline that uses a purely geometry-based prediction and association~\cite{yin2021center}, our framework improves the main metric \ac{amota} substantially by $\SI{5.2}{\percent}$. The optimized version of MUTR3D~\cite{zhang2022mutr3d, mutr3d_eval_bug} is outperformed by $\SI{1.9}{\percent}$, highlighting the crucial role of the \ac{lmm}. 
In particular, we observe a drastic reduction of \ac{ids} by $\SI{86.2}{\percent}$ compared to the greedy version and by $\SI{28.7}{\percent}$ compared to MUTR3D, see~\tabref{table:nusc_results}. We address this fact to the spatially and temporally consistent appearance representations provided by the \ac{lmm} and our proposed track embedding. This benefits the association resulting in less track fragmentations~(\ac{frag}) and a higher amount of mostly tracked trajectories~(\ac{mt}).

Additionally, \methodName\ also outperforms the concurrent works PF-Track~\cite{pang2023standing} by $\SI{1.7}{\percent}$ and CC-3DT~\cite{fischer2022cc} by $\SI{2}{\percent}$ \ac{amota}, respectively. The former employs advanced query refinement operations for temporal consistency and a stronger VovNet-V2-99~\cite{lee2019energy} image backbone, the latter proposes a LSTM-based learned motion model~\cite{fischer2022cc}. 
Evaluating our model with a VovNet-V2-99 trained on both the train and validation set on the nuScenes test set results in $\SI{43.9}{\percent}$ \ac{amota}. This improves over MUTR3D~\cite{zhang2022mutr3d} by $\SI{16.9}{\percent}$ and even outperforms concurrent work that utilizes stronger detection algorithms~\cite{fischer2022cc, pang2023standing}.

\subsection{Ablation and Analysis}
\begin{figure*}[t]
  \includegraphics[width=\textwidth]{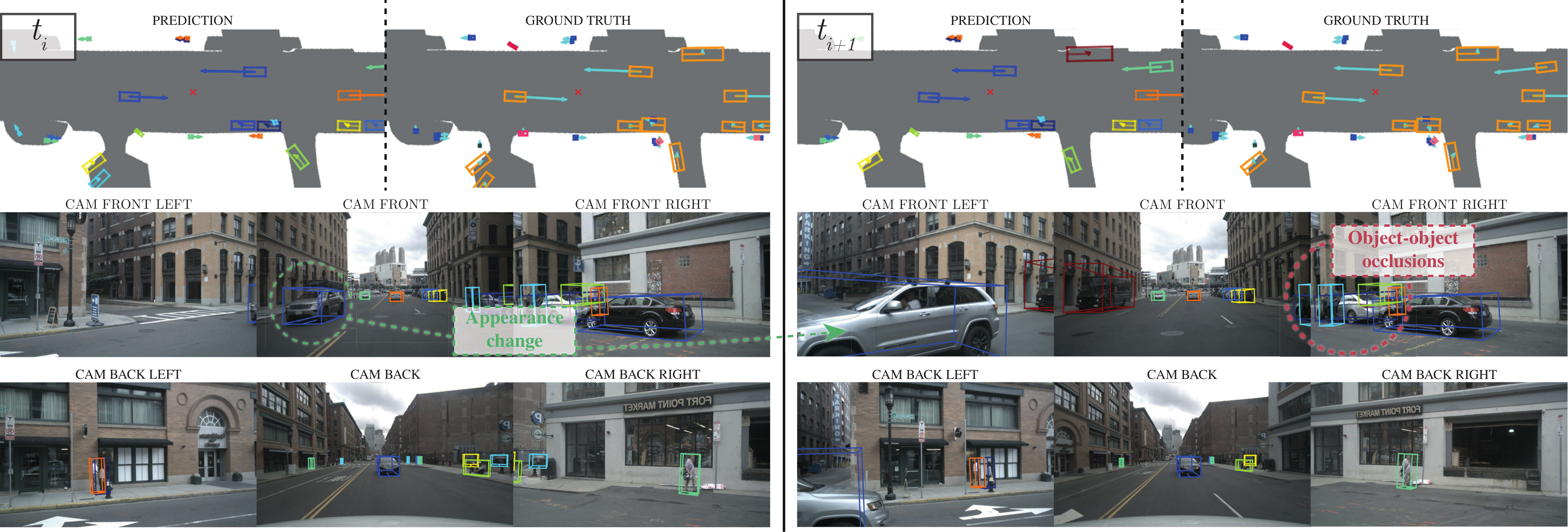}
  \caption{Qualitative results for two consecutive frames on the nuScenes~\cite{caesar2020nuscenes} validation set. Upper row shows predictions and ground truth in top view. Different colors of the predicted objects indicate different object ids. The bottom row shows the predictions projected to the multi-view camera images.}
  \label{fig:qualitative_results}
\end{figure*}

\boldparagraph{Qualitative results}
A qualitative example of two consecutive time steps is shown in \figref{fig:qualitative_results}. \methodName\ is particularly strong in handling large appearance changes, e.g. due to different lighting conditions and tracking road participants under strong object-object occlusions. We provide additional videos of the tracking performance in the supplementary.

\boldparagraph{Effect of Training Time}
The effect of longer training schedules is shown in~\tabref{table:training_shedule}. MUTR3D~\cite{zhang2022mutr3d} gains a performance boost of $\SI{2.0}{\percent}$ in \ac{amota} and $\SI{7.3}{\percent}$ in \ac{ids} by further fine-tuning. Adding the proposed \ac{lmm} yields $\SI{2}{\percent}$ \ac{amota} and improves the \ac{ids} by $\SI{10.7}{\percent}$ as compared to the equally long trained model. This clearly indicates that the use of our \ac{lmm} results in more consistent tracks with a reduced number of identity switches.

\begin{table}
  \begin{center}
    \caption{Effect of training time. For a fair comparison we fine-tune our version of MUTR3D~\cite{zhang2022mutr3d} with and without an \ac{lmm} indicated by w/\ac{lmm}. Runs denoted by w/Init use a pretrained MUTR3D instead of a pretrained DETR3D~\cite{wang2022detr3d} checkpoint.} 
    \label{table:training_shedule}
    \begin{tabular}{ccccc}
      \hline\noalign{\smallskip}
      w/\ac{lmm} & w/Init & AMOTA$\uparrow$ & AMOTP$\downarrow$ & IDS$\downarrow$\\
      \noalign{\smallskip}
      \hline
      \xmark & \xmark & 0.338 & 1.425 & 531\\
      \xmark & \cmark & 0.358 & 1.382 & 492\\      
      \cmark & \cmark & \textbf{0.378} & \textbf{1.365} & \textbf{439} \\
      \hline
    \end{tabular}
  \end{center}
\end{table}

\begin{table}
  \begin{center}
    \caption{Effect of different \ac{lmm} architectures.  w/\ac{lmm} indicates whether an \ac{lmm} is used, multi-head (w/MH) denotes a sparse latent transformation matrix $\tfl{b}{a}$ instead of a full-rank version. The head / matrix size is denoted by $\mid K \mid$.}
    \label{table:lmm_architecture}
    \begin{tabular}{lcccc}
      \hline\noalign{\smallskip}
      w/\ac{lmm} & w/MH & $\mid K \mid$ &AMOTA$\uparrow$  & IDS$\downarrow$\\
      \noalign{\smallskip}
      \hline
      \xmark & - & - & 0.358 & 492\\
      \cmark & \xmark & $32^2$ & 0.372 & 432\\
      \cmark & \xmark & $96^2$ & 0.370 & \textbf{402}\\
      \cmark & \cmark & $16 \cdot 16^2$ & 0.374 & 517\\
      \cmark & \cmark & $4 \cdot 64^2$ & 0.373 & 434\\
      \cmark & \cmark & $8 \cdot 32^2$ & \textbf{0.378} & 439 \\
      \hline
    \end{tabular}
  \end{center}
\end{table}

\boldparagraph{Effect of \ac{lmm} Architecture} 
The performance of different \ac{lmm} architectures is shown in~\tabref{table:lmm_architecture}. Using the proposed sparse multi-head \ac{lmm} instead of a full-rank representation of the latent motion matrix $\tfl{b}{a}$ does not only align the architecture to the multi-head attention blocks but also reduces the amount of output parameters of the hyper-network. This is key to scale the latent transformation matrix to the full latent space dimensions. Using the same configuration as the attention blocks of the transformer for the multi-head \ac{lmm} results in an boost in \ac{amota} of $\SI{0.8}{\percent}$ over a \ac{lmm} that uses a full-rank latent motion matrix.

\boldparagraph{Effect of Transform Representation}
Different strategies to apply the transformation modeled by the \ac{lmm} are shown in \tabref{table:lmm_setup}. We do not observe a performance increase when the latent query is used as an additional input to the \ac{tfnet}. This is in line with our general design paradigm to compute the latent motion matrix solely from its geometric counterpart. Although it is beneficial to apply the \ac{lmm} twice instead of merging object and ego motion, using shared parameters for the object and ego motion compensation cuts the number of parameters in half and does not cause any ill-effects. This supports the general design to model any geometric transformation with the \ac{lmm} without creating an explicit distinction between object and ego motion.

\boldparagraph{Integration to other methods}
To showcase the flexibility of the proposed \ac{lmm} we incorporate it into the concurrent work StreamPETR~\cite{wang2023exploring} that proposes \ac{mln}. As shown in \tabref{table:stream_petr} our proposed architecture improves the NDS by $\SI{1.3}{\percent}$ when using the \ac{lmm} which is a generalized version of the \ac{mln}.

\boldparagraph{Inference latency}
An analysis of the runtime of different components of \methodName\ is shown in~\tabref{table:runtime}. The proposed \ac{lmm} only adds additional $\SI{1.48}{\percent}$ latency and the track-embeddings $\SI{1.52}{\percent}$ respectively, since the runtime is dominated by the image backbone and transformer layers for both DETR3D~\cite{wang2022detr3d} and StreamPETR-based~\cite{wang2023exploring} models.

\begin{table}
  \begin{center}
    \caption{\ac{lmm} transform representation. Models that apply object and ego motion separately are denoted with w/Separate. w/Share indicates models that use shared parameters and w/Feats \ac{lmm}s that utilize the query feature as input to the \ac{tfnet}.} 
    \label{table:lmm_setup}
    \begin{tabular}{ccccc}
      \hline\noalign{\smallskip}
      w/Separate & w/Share & w/Feats & AMOTA$\uparrow$ & IDS$\downarrow$\\
      \noalign{\smallskip}
      \hline
      \xmark & \cmark & \xmark & 0.370 & 492\\
      \xmark & \cmark & \cmark & 0.371 & 446\\
      \cmark & \xmark & \xmark & 0.377 & \textbf{411}\\
      \cmark & \xmark & \cmark & 0.366 & 464\\
      \cmark & \cmark & \cmark & 0.370 & 426\\
      \cmark & \cmark & \xmark & \textbf{0.378} & 439\\
      \hline
    \end{tabular}
  \end{center}
\end{table}

\begin{table}
    \begin{center}
    \caption{Performance of StreamPETR~\cite{wang2023exploring} on the validation set of the nuScnees detection benchmark. $\mathparagraph$ indicates a model that uses the proposed \ac{lmm} instead of the \ac{mln}.}
    \label{table:stream_petr}
    \begin{tabular}{lcccccc}
        \hline\noalign{\smallskip}
        Name & mAP$\uparrow$ & mATE$\downarrow$  & mAOE$\downarrow$ & mAVE$\downarrow$ & NDS$\uparrow$\\
        \hline
        StreamPETR & 0.483 & \textbf{0.591} & 0.479 &  0.195 & 0.562\\
        StreamPETR$\mathparagraph$ & \textbf{0.485} & 0.611  & \textbf{0.367} & \textbf{0.185} & \textbf{0.575}\\
        \hline
    \end{tabular}
    \end{center}
\end{table}

\begin{table}
    \begin{center}
    \caption{Latency of different components of the proposed tracking framework in milliseconds on a NVIDIA A100 GPU. $*$ shows a version without track-embeddings, $\mathparagraph$ uses StreamPETR~\cite{wang2023exploring} with a \ac{lmm} instead of DETR3D as detection transformer.}
    \label{table:runtime}
    \begin{tabular}{lc|cccc}
        \hline\noalign{\smallskip}
        Name & Total & Backbone & Transformer & LMM\\
        \hline
        \methodName* & 145.6 & 33.7 & 80.1 & $\mathbf{2.2}$\\
        \methodName & 147.8 &  33.7 & 80.1 & 4.4\\
        StreamPETR$\mathparagraph$ & $\mathbf{49.7}$ &  $\mathbf{19.1}$ &  $\mathbf{13.6}$ & 3.3\\
        \hline
    \end{tabular}
    \end{center}
\end{table}
\section{Conclusion}
This paper presented \methodName, a novel approach for 3D object tracking-by-attention that is compatible with any query-based object detector. We transferred the concept of motion models from traditional geometry-based trackers to the tracking-by-attention paradigm in terms of latent motion models that predict the spatio-temporal appearance change of objects between two frames. This allowed for a prediction step that models a geometric transformation in an analytical way and applies this transformation in the latent space with a learned motion matrix at the same time. An additional latent track embedding improved the latent existence probability of tracks. In our experimental evaluation, the integrated system demonstrated significant improvements in all relevant tracking metrics. Increased track consistency was observed as a particular strength evident from significantly decreased identity switches and track fragmentations. 

We hope that this work serves as a foundation for future tracking-by-attention research with the aim of integrating model-based assumptions to end-to-end tracking approaches. While the potential of this has been clearly demonstrated in this work, limitations and opportunities for improvement have also been identified.

\boldparagraph{Limitations}
The implicit association used in the tracking-by-attention scheme falls short in cases with poor motion estimates, since the resulting prediction might impair the re-identification performance in the next frame. This could lead to errors in object position or track losses. In future work, multi-hypothesis tracking~\cite{kim2015multiple} could be adopted to model uncertainty in object dynamics and to relax the one-to-one relation of track queries between frames. Additionally, the implicit assignment results in a discrepancy between training and inference time, since the ground truth matching only assigns the correct ground truth object to a single query during training. This could be solved with a non-strict matching approach as demonstrated in 2D tracking~\cite{jia2023detrs}. The novel idea of track embeddings is a promising research direction that could be extended to model the uncertainty distribution of each tracked object explicitly.
\flushend

% \bibliography{IEEEabrv,bibliography_long,bibliography}
%
% \bibliographystyle{ieee_fullname}
\bibliographystyle{IEEEtran}
\bibliography{bibliography}

\end{document}